# OFF-LINE ARABIC HANDWRITTEN WORDS SEGMENTATION USING MORPHOLOGICAL OPERATORS

Nisreen AbdAllah[1] and Serestina Viriri[2]

[1]Sudan University of Science and Technology, Sudan
[2]University of KwaZulu-Natal, South Africa

*ABSTRACT*

*The main aim of this study is the assessment and discussion of a model for hand-written Arabic through segmentation. The framework is proposed based on three steps: pre-processing, segmentation, and evaluation. In the pre-processing step, morphological operators are applied for Connecting Gaps (CGs) in written words. Gaps happen when pen lifting-off during writing, scanning documents, or while converting images to binary type. In the segmentation step, first removed the small diacritics then bounded a connected component to segment offline words. Huge data was utilized in the proposed model for applying a variety of handwriting styles so that to be more compatible with real-life applications. Consequently, on the automatic evaluation stage, selected randomly 1,131 images from the IESK-ArDB database, and then segmented into sub-words. After small gaps been connected, the model performance evaluation had been reached 88% against the standard ground truth of the database. The proposed model achieved the highest accuracy when compared with the related works.*

## 1. INTRODUCTION

Handwriting recognition has attracted the researcher's attention in Optical Character Recognition (OCR) area and needs more integration efforts between different research so that to reach satisfy outcomes [1]. Handwriting recognition simulates human writing in symbolic representation [2], [3] which has been an intense topic in the past three decades. Online and offline are types of handwriting recognition systems. The online type was made at the time of writing and offline after the writing was completed [4].

Handwriting cursive recognition is very challenging [5], also, the Arabic language has a special situation as the overlap between characters and the presence of diacritics like dots and Hamza complicated the task [6]. This is task seems simple, but it isn't. Even for human beings when the absence of full context [7]. Day to day services are offered by offline handwriting recognition, which can be summarized in forms processing, archiving books, signature, writer identification, bank check transaction, and documents editing.

In North Africa and Middle East areas Arabic is the main language. Arabic, Farsi, Sindhi, Pashto, and Urdu in most of the alphabets being the same in writing [8]. Arabic words written from right to left, character after character without spaces in between. However, six characters are linked from the right and disjoint from the left [1]. Samples of characters disjoining words. Characters are written in hand and print, with their names in Arabic and English, are shown in Figure 1.

| Handwriting | ا | ر | ز | د | ذ | و |
|---|---|---|---|---|---|---|
| Printed | ا | ر | ز | د | ذ | و |
| Ch.Name Arabic | الف | راء | زاي | دال | ذي | واو |
| Ch.Name English | Alif | Raa | Zaay | Daal | Dhaal | waaw |

Figure 1. Characters disjoining words

The Arabic words are composed of 28 characters. Each character has 2 to 5 shapes each in a particular position, moreover, is linked in a cursive form in a single stroke, and that why segmentation is seen as a difficult task [9].

The Arabic Alphabet contains 15 characters hold dots. Dots position above or below the primary character body. Just 6 among 28 characters are connected from the right side when are located in the middle. However, other characters connected into both sides left and right [10].

To enumerate sub-words in the word depends on the number of the characters disjoining. This form causes the sub-words phenomenon and also known as (PAW) Pieces of Arabic Words in Arabic and Arabic-like languages [1][11].

The primary body of the word is completely connected if it doesn't have a disjointed character. Otherwise, the word is partially connected if containing more than one disjoined character. Two samples of handwriting words, completely or partially connected, dots positions and sub-word numbers are shown in Figure 2 and Figure 3.

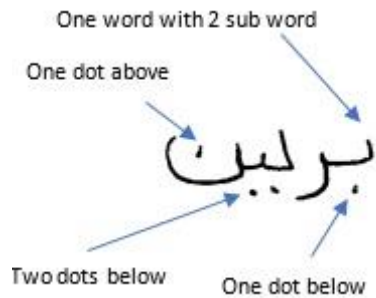

Figure 2: "Barleen" Arabic word composed of two sub-words

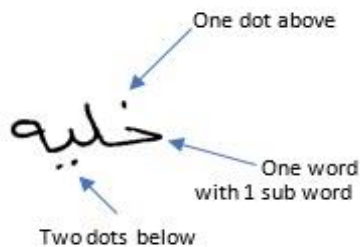

Figure 3: "khalyia" Arabic word composed of one sub-word.

Cursive word is broken at the moment of pen lifting-off or while scanned documents then lost slightly parts of the word. The previous case leads to holes or gaps in the cursive word. The holes lead to a disconnect in the word body. Most Arabic handwriting segmentation errors are caused by gaps or discontinuity [12] [13].

The OCR has four stages: pre-process, segmentation, feature extraction, and classification. The result of a reliable and efficient OCR system depends on the first pre-processing essential stage [14] and incorrect parts may lead to invalid classification or refuse characters [15].

In conclusion, Arabic is written in a cursive form. Six characters disjoining from the left side [16]. Gaps in writing are produced due to lifting-off the pen, ink fades, and whiles after applying pre-processing operations [17].

Handwriting recognition faces many challenges different from print. The challenges of handwriting, as shown in Figure 4, are variant in styles, thickness changes, different writing materials, ligature, touching, and overlapping between characters. The previous challenges are not found in printed styles because the words have the same styles, thickness, and do not have ligature in most font types.

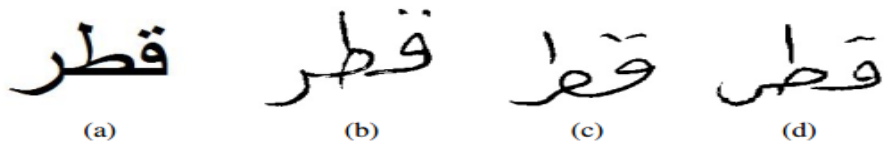

Figure 4. (a) Printed word, (b), (c), and (d) The same word was hand written in different styles.

The researchers face many challenges in Arabic handwriting recognition discussed with suggested solutions in more detail in [1][18][19].

This paper presents three main ideas. The first idea is to design a pre-processing method to connect gaps CGs using a combination of three morphological operators. The second idea is to implement a model to segment words into sub-words. The third idea is to evaluate the result of segmentation against the ground truth.

The remaining parts of this paper are arranged as follows: Section 2 describes the related works. Section 3 shows and represents the methods and techniques implemented in the proposed model stages. Section 4 discusses the experimental results. Finally, the conclusion and future work discourses in Section 5.

## 2. RELATED WORKS

Many research has been carried out in the field of segmentation of offline Arabic handwritten scripts; however, still requiring intensive studies to segment word into characters. There are two approaches sued in recognition systems: segmentation-based and segmentation-free. Segmentation-based is an approach to segment words into characters or small units. This type also knows as an analytical approach. While segmentation free takes the word as a whole unit. This type knows as a global approach.

Some, when used the first approach, assumed there were no gaps while writing the words, even if it already occurred [20]. Most studies prefer using the second approach and deal with a word as a whole and complete unit due to the variability and unconstrained of human writing [21].

This paper tries to segment words into sub-words using the first approach, segmentation-based. Our motivation is that there exist a few efforts to segment words into subwords. The two subsections below discuss the related works in morphological operation and connected component technique then summarize that in Table 1.

## 2.1. Morphological Operators

Morphological operators are often applied as a filter to reduce or remove noise from images in the pre-processing stage. Filters enhance and improve OCR system results. Slight studies have been investigated for Arabic character segmentation based on morphological operators [15].

A review was introduced using sample methods designed to remove noise that might appear through scanned documents images discussed intensively in [22]. Therefore, research in this area demands to be covered widely.

Most research in Arabic handwriting has been done in the pre-processing stage and focused on baseline detection and skew correction as in [23],[24],[25]. A deep discussion for mathematical morphology algorithms that enhance images performing before recognition discoursed in [26].

Due to the importance of pdf documents in day-to-day services, much research has been done. One of these recent research conducted by NH Barna et al. [27], that analyzed various components of documents using opening and dilation and other morphological operations. The method was segmented printed documents into text, image, table, and cell using the bounding box technique. Different sources for this research originate from digitally online and manually pdf scanned data. The accuracy when calculated illustrate table and cell have the best than image and text.

On the other side, some research recently used two morphological operations the top hat and the bottom hat as a filter to remove unwanted noise and contrast enhancement of the medical, remote sensing, and natural images [28].

An investigation was done and carried out using six morphological operators proposed as in [29] to enhance binary images extracted from twenty-five shapes. The six morphological operators were applied to remove the noise without modifying the shape. These operators were dilation, erosion, opening, closing, fil, and majority. Experiment results are subjectively discussed when two or more operators are combined, they significantly enhance images.

Motawa et. Al [30] merged morphological operators and connected components to segment words into characters after detect and correct slant stroke. The algorithm went through three steps: the filter of closing followed by opening, singularities, and regularities. Also, morphological operation filter can generate temporary to fix bit gaps in words while applying methods for character segmentation as in [14].

## 2.2. Connected Compounds CCs

Measuring distances between connected components in Arabic words, AlKhateeb et al. introduced a method using a vertical histogram and connected components to segment sub-words. The technique analyzed the line to decide if space areas correspond to one word or two. Then applied on 200 images from IFN/ENIT database and achieved an accuracy result of 85% [31].

To resolve the sub-words overlapping issue in Arabic handwritten script, Ghaleb et. al. proposed the pushing technique, which includes connected component labeling and threshold to get a clear vertical segmentation. This method evaluation was using IESK-ArDB, KHATT, and IFN/ENIT as an Arabic example and a Persian database and obtained 72% as an accuracy result [13].

Extra technique analyzed sub-word distance. Distance between bounding boxes was used and divided into main and auxiliary connected components to determine cutting points between boxes. Furthermore, they extract features to separate the connected component. The algorithm was tested on 450 images from IESK-ArDB without declaring the overall segmentation accuracy [32].

Table 1. Summarized of related works.

| # Ref | Year | Database | ImagesNum | Techniques | Seg.Type |
|-------|------|----------|-----------|------------|----------|
| [29] | 2008 | Private | 25 | 6 Operators | Pattern |
| [30] | 1997 | Private | Few hundred | Closing-Opening | One word |
| [12] | 2012 | IFN/ENIT | 1250 | Morphological filter | Sub-words |
| [31] | 2009 | IFN/ENIT | 200 | CCs | Sub-words |
| [13] | 2015 | 4 Databases | 400 | CCs & Threshold | Sub-words |
| [32] | 2016 | IESK-ArDB | 450 | BBox distance analysis | Sub-words |

## 3. METHODS AND TECHNIQUES

The framework of the model was composed of three: pre-processing, segmentation, and evaluation stages are shown in Figure 5. The objectives of the model were enhancing and filtering binary images that contained disconnected gaps and segmenting the word into sub-words. The methodology is based on assumption that gaps were in the words without separation into a new group.

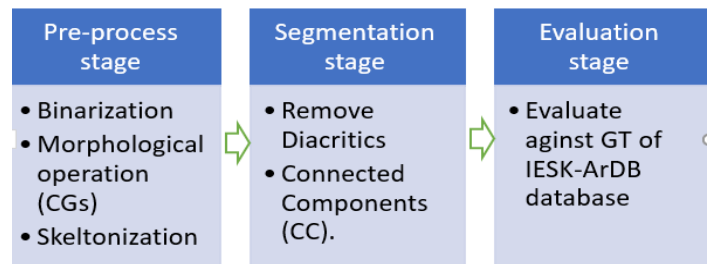

Figure 5: The proposed model framework

### 3.1. Select a Database

The most effective factors to produce real-life handwriting recognition systems for the marketplace are availability and the unconstraint of a huge database. Moreover, result evaluating using a benchmark through standard ground truth.

Many studies came out in Arabic handwritten segmentation words. However, datasets were collecting locally and evaluated privately, accordingly, it was inapplicable to compare results from different databases. Although the IFN/ENIT database is familiar, available, and has not a ground truth clarify characters or sub-words position.

Thus, the IFN/ENIT database did not have segmentation-based process requirements. Despite this situation in [12] using this database by adding a manual file to evaluate segmentation results.

Signal & Image Processing: An International Journal (SIPIJ) Vol.11, No.6, December 2020

Therefore, to automatically evaluate the proposed model of CGs, the IESKArDB database [33]was voted and elected.

## 3.2. Model Algorithm

This section introduces the proposed CGs model. The model was implemented using two stages pre-processing and segmentation to resolve the gap in Arabic handwriting words, then evaluate the model against ground truth. Connect Gap model steps of Arabic handwriting manifest shown in Figure 6.

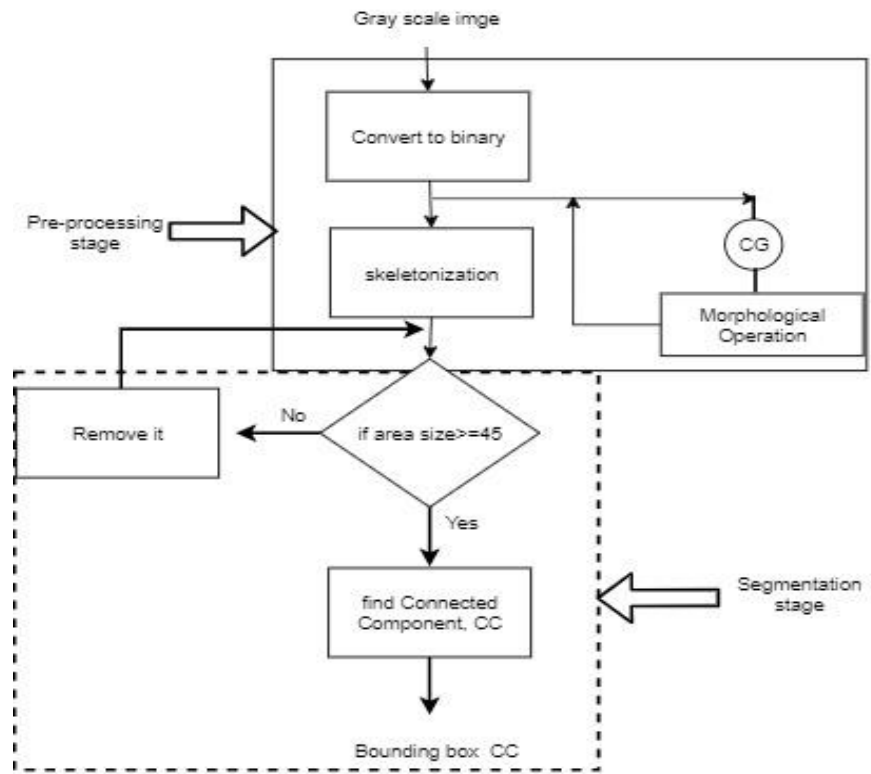

Figure 6: Steps followed in Connected components model (CGs)

### 3.2.1. Pre-processing stage

A pre-processing operation is the first stage applied in OCR systems. The main objective of pre-processing in an OCR system is to prepare and enhance the images to the next stages. Before segmenting handwriting images or recognition by the machine, it requires many processes to enhance and prepare images such as filtering and removing noise. Maybe, the noises were due to low-quality paper, ink fade, and irregular hand movement [34]. In the model pre-processing stage contains binarization, morphological operation, and skeletonization steps. The next subsections discuss these steps in more detail.

#### 3.2.1.1. Binarization

Is used to generate a binary image with zeros and ones depending on the threshold values. The images in the database stored in grayscale; hence, Otsu's [35] was applied. Otsu's utilized threshold value automatically and minimize variation in images.

26

Signal & Image Processing: An International Journal (SIPIJ) Vol.11, No.6, December 2020**3.2.1.2. Morphological operators**

Produced a new image after modifying binary image pixel-by-pixel using mathematical operators. In the algorithm shown in Figure 7, the model requires a binary image before applying morphological operators to connect gaps. Three operators merge to bridge the gaps in Arabic words. These operators were described as follow:

1. Add pixels to expand the outer edge of the word. Repeat the process for the outer edge four times.
2. Set 0-valued pixel to 1, if it has two neighbors that have a 1-valued. Repeat the process for the outer edge twice.
3. Test 8-connectivity of the pixel, if five or more of them have 1-valued set the pixel to 1. Otherwise, set the pixel to 0. Repeat the process for the outer edge twice.

**3.2.1.3. Skeletonization**

Commonly known as thinning, it is an important and crucial step in the OCR application. Skeleton step is used to reduce a binary image to 1-pixel using 4-connectivity and preserve image basic structure. The CGs model used a fast skeleton which was implemented in [36] and might effectively extract word skeleton.

**3.2.2. Segmentation Stage**

Segmentation is a process that attempts to decompose an image into subunits in an intelligent process after analyzing the content [37]. It is not an easy process, moreover, it depends upon various writing styles. Connected component (CC) is one of the strategies to segment an image by using a bounding box analysis [38]. The CC technique is used to bound the continued component into boxes. The CC technique was improved in [39] and [40], then introduced in a simple and efficient design in [41] which was used to label the CGs algorithm.

```
Algorithm 1 CGs Algorithm
    Input: w-img(wordImage), t(threshold)
    Output: x-img
 1: procedure CONNECT GAPS(w-img,t)
 2:     b-img ←CreateBin(w-img,t)        ▷ Create Binary Image
 3:     x-img ←IntBinImage(b-img)
 4:     for all outerBoundary(w-img), i = 1, i++ do
 5:         for each pixel(p) belong to w-img do
 6:             while counter ≠ 4 do
 7:                 x-img ←expImg(w-img,8)     ▷ Edge Expansion
 8:                 n = numNonZeroNeighbors(p)
 9:                 if x-img[p] == 0 then
10:                     if n == 2 —— n ≥ 5 then
11:                         x-img[p] = 1
12:                     end if
13:                 end if
14:             end while
15:         end for
16:     end for
17:     return x-img
18: end procedure
```

Figure7: CGs Algorithm

27

Signal & Image Processing: An International Journal (SIPIJ) Vol.11, No.6, December 2020

## 4. EXPERIMENTAL RESULTS AND DISCUSSION

Several experiments were examined in this section to evaluate the proposed model. The available and free database was used to validate the model. The proposed model results were compared with existing algorithms. The results were clarified in the following table.

Table 2. Segmentation evaluation (%) of the model.

| Ref No# | Accuracy | Precision | Recall | Specificity | F-score |
|---|---|---|---|---|---|
| **Proposed Model** | **88** | **89** | **99** | **13** | **93** |
| [29] | Made as Remarks | | | | |
| [30] | 81.88 | Not counted | | | |
| [12] | 70 | Not counted | | | |
| [31] | 85 | Not counted | | | |
| [13] | 72 | Not counted | | | |
| [32] | Not Declared | Not counted | | | |

The general accurate segmentation of the model was illustrated in Table 2. The proposed model obtained 88% as an accuracy result and 12% as a segmentation error rate.

The model segmentation errors were occurred due to variation in writing style, especially too long space of lifting-off the pen as showed in Figure 8 and Figure 9 while separated component found to lead to errors and influences the segmentation result. On the other hand, the touching component closed spaces lead to segmentation errors.

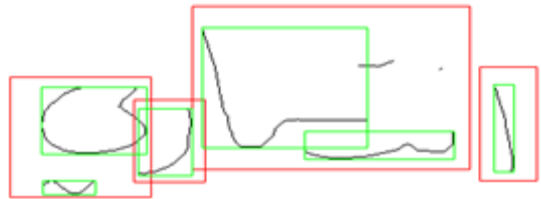

Figure 8: Over Segmentation due to lifting-off for long spaces

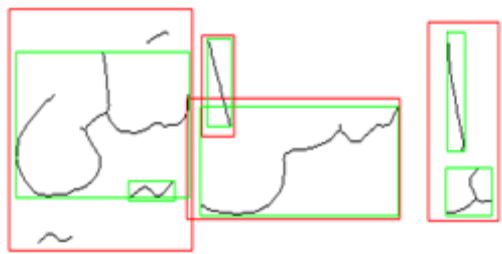

Figure 9: Over segmentation due to the large size of dots and Hamza

As one of the obvious limitations, the results of the current model found that handwriting of Arabic words especially dots and Hamza create more errors in segmentation due to variation in handwriting styles Figure 8 and Figure 9.

the proposed model expanded Arabic words which can lead to touching close parts of the letters. As well as, small parts like dots and Hamza can be bigger which leads to misclassification. For that, the CGs model could not suitable for document words segmentation, which might word touch each other.

Signal & Image Processing: An International Journal (SIPIJ) Vol.11, No.6, December 2020

## 4.1. Database

The available and free database was used in the segmentation stage. The proposed model automatically evaluated using ground truth files attached to the IESK-ArDB database. The below section describes a particular database and its contents.

The IESK-ArDB dataset contains 4,000 words images with 350 dpi resolution in various sizes. The forms were designed using 8 pages and each page was filled with eight handwritten words. The dataset was collected using 22 writers from several Arabic countries. Some greyscale samples of the dataset shown in Figure 10. The database considered the distribution of basic Arabic letters in various positions (Begin, End, Isolated, Middle). More descriptions for this database found in [23].

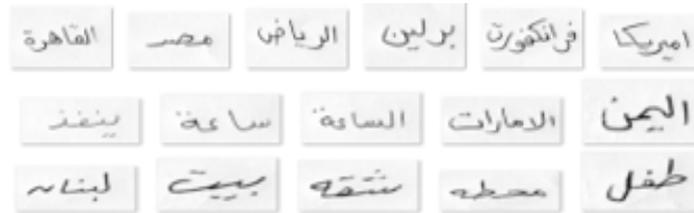

Figure 10: IESK-ArDB database samples

The database was already divided into 12 parts. Each part kept both inputs as grayscale images in BMP format and ground truth information in XML format. Only one image identified by the ID Q01-006 did not get complete information in the ground truth. The ground truth contained ID, LetterLabel, baseline, sub-words, and more detailed information such as age and gender. Sub-words tags contained pixels (ax, ay) and (bx, by), to establish upper-left and bottom-right bound coordinates. The LetterLabel tag contained the name, shape, and Unicode for each character in the word.

Selected 1,131 images randomly from IESKArDB to test and evaluate the proposed model. A word in the database may contain one or more sub-words. The chosen images included 2652 sub-words. Most Arabic words in the database that will be used in the experiment were contained in more than one piece. Words that were selected to experiment proposed model were composed of variant categories as histogram shown in Figure 11.

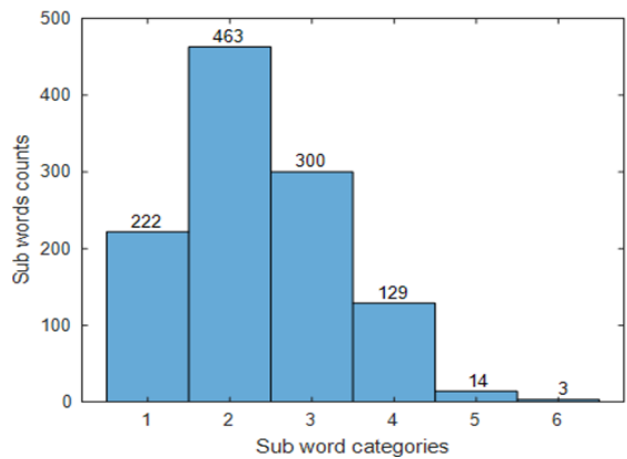

Figure 11: Sub-words Histogram categories

## 4.2. Model Implementation

The proposed model was implemented using MATLAB R2018a version, windows 10 pro 64-bit Operating System, with RAM 6.00 GB, CPU 1.80 GHz core i5.

## 4.3. Calculation Metrics

The most common image segmentation evaluation metrics to compare results such as Accuracy, precision, recall, f-score, and specificity. Metrics that were used for model evaluation were discussed in the below section.

1- Accuracy:

Measures the ratio of true positive classes and true negative classes overall classes examined. It means how often is the classifier correct.

$$Accuracy = \frac{(TP + TN)}{TP + FP + FN + TN}$$

2- Precision:

It measures the ratio of the true positive classes over all the predicted positive classes.

$$Precision = \frac{TP}{(TP + FP)}$$

3- Recall:

It measures the ratio of true positive overall actual positive classes. The recall is also known as sensitivity.

$$Recall = \frac{TP}{(TP + FN)}$$

4- Specificity:

It measures the ratio of true negative overall actual negative classes.

$$Specificity = \frac{TP}{(TN + FP)}$$

5- F-score:

This measure evaluates how the object boundary matched the ground truth boundary.

$$f - score = \frac{2 * Precision * Recall}{(Recall + Precision)}$$

Where TP: true positive, TN: true negative, FP: false positive, and FN: false negative.

Signal & Image Processing: An International Journal (SIPIJ) Vol.11, No.6, December 2020

## 4.4. Results and Discussion

The proposed model was evaluated on 1,131 images from IESK-ArDB. The segment sub-words part 1 to 5 were chosen to test the model. Three steps were applied to bridge the gap. In this experiment, some sources of error were observed. The main of these errors was touching between contiguous sub-words in the same word after applying expanded operators.

The proposed model utilized a huge data to apply a variety of handwriting styles so that to be more compatible with real-life applications and get real achievements.

The proposed model was compared with the findings of the related method in the literature and achieved an accuracy of 88% as the best result. In addition to that, the proposed model was evaluated using huge data comparing with other related works. As well, was evaluated automatically using the standard ground truth of the database. On the other hand, this result shows that the proposed model can connect small gaps and segment these words into its subwords properly.

The bounding box detection technique is used to evaluate and test the proposed model. A case of evaluations is shown in Figure 12, the proposed model against ground truth boxes. The red box represents the ground truth, and the green box represents the proposed model. Three cases of evaluation are excellent, good, poor. As an example, while evaluated the connected components, sub-words bounded using a minimal bounding box surrounding the word exactly after removing the dots. Nevertheless, the ground truth bounded the character and the dots as one word using maximal boxes, and some handwriting styles write dots far from characters Figure 13, which leads to minimizing the overlap ratio.

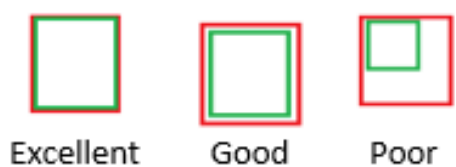

Figure 12: Three cases of bounding box Detection evaluation

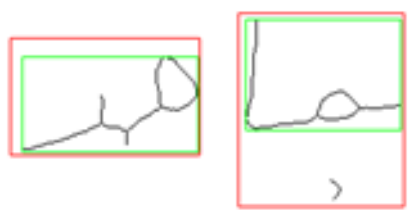

Figure 13: The dot far from the sub-word body; minimum overlap ratio.

The model was unable to detect sub-words which were 13% of all sub-words total. Undetected sub-words as shown in Figure 14, which are called specificity, had been appeared due to the size less than 30 was lost or dropped. The proposed model removed a small size less than 30 which may contain be dots, Hamza, or other small characters. The touching between sub-words leads to under-segmentation as shown in Figure 14 and Figure 15. Touching between sub-words leads to under-segmentation. Under-segmentation occurred when the number of boxes was less than the ground truth.

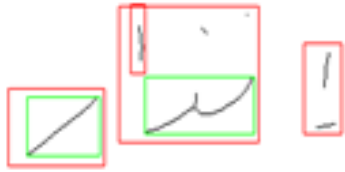

Figure 14: Size less than 30; losing parts.

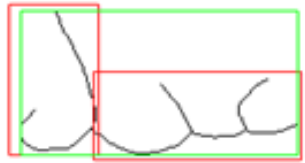

Figure 15: Under-segmentation; touching two sub-words.

Some examples of these results are shown in Figure 18, Figure 20, and Figure 22 as samples of gaps in words while pen lifting-off or during documents scanned. After applying CGs, gaps were connected as presented in Figure 19, Figure 21, and Figure 23. The connected gap led to true segmentation shown in Figure 16 and Figure 17. The proposed model boxes had been equals to numbers of the ground truth.

However, some words had a long space the CGs unable to connect them, as shown in Figure 24, and Figure 25, which led to over-segmentation as shown in Figure 26 Figure 27. The over-segmentation has occurred when the number of boxes in the proposed model greater than the number of the ground truth.

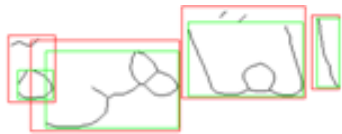

Figure 16: One word contains 4 sub-words; true segmentation.

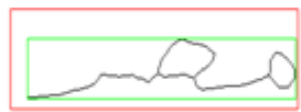

Figure 17: One word contains 1 sub-word; True segmentation.

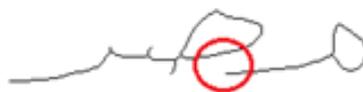

Figure 18: A circle shows the gap position before CGs.

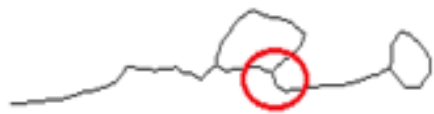

Figure 19: A circle shows the gap connect after CGs.

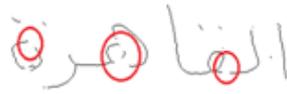

Figure 20: Circles show gap positions before CGs.

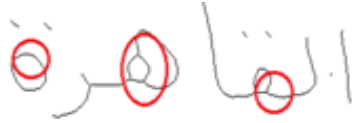

Figure 21: Circles show gaps connect after CGs.

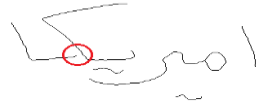

Figure 22: A circle shows the gap position before CGs.

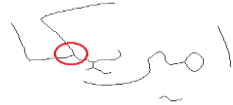

Figure 23: The circle shows the gap connect after CGs.

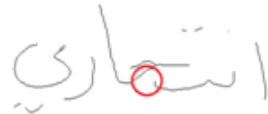

Figure 24: CGs was not successful to connect the gap; too long horizontal space.

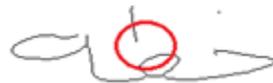

Figure 25: CGs was not successful to connect the gap; too long vertical space.

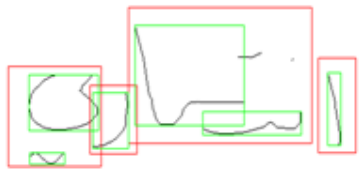

Figure 26: Over-segmentation; long space and large dots.

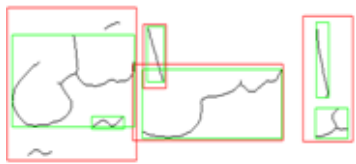

Figure 27: Over-segmentation; the large size of Hamza and dots.

## 5. CONCLUSIONS

The Arabic hand-writing model based on segmentation was well presented. Pre-processing steps were essentially applied to convert images into a binary mode. It also connects the small gaps to make images ready for segmentation. Subsequently, the connected component technique was used to segment words into sub-words using bounding boxes. Ground truth of IESK-ArDB database files compared with boundary boxes to evaluate the model. Pre-processing transformation enhanced the segmentation results.

In future work, the model should develop by investigating and resolving handwriting issues such as overlapping, touching, and ligature in sub-words. Resolving these issues intensive investigated research needs to be done in this area by improving more pre-processing filters to connect long gaps. As a suggestion to improve the proposed model can be integrated using the seam carving algorithm in [42]to resolve the touching problem. In addition to that, resolving ligature can apply the techniques using [43] and [44]. Arabic handwriting segmentation requires extensive research to produce suitable solutions to segment connected components to the characters correctly for producing handwriting recognition used in real-life services. Future trends of Arabic character recognition can be found in this survey in more detail [45].

## CONFLICTS OF INTEREST

The authors declare no conflict of interest.

## ACKNOWLEDGMENTS

The authors would like to thank Laslo Dings and his colleagues for their efforts and assistance to access the full database through the internet.

## AUTHORS

**NISREEN ABDALLAH** received a BSc degree in computer science from Omdurman Islamic University, and an M.Sc. degree from the University of Khartoum, Sudan. She is currently pursuing a Ph.D. degree in computer science at Sudan University of Science and Technology, Sudan. She also works as a lecture in the Department of Information Technology at the University of Kordofan. Her research interests include machine learning, computer vision, image processing, and pattern recognition. She attended and presented in some national and international conferences about handwriting recognition.

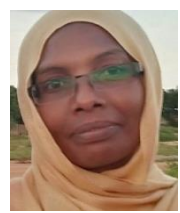

**SERESTINA VIRIRI** (Member, IEEE) is currently a professor of computer science. He is also working as a Professor of the school of mathematics, Statistics, and Computer Science, University of Kwazulu-Natal, South Africa. He is an NRF-rated researcher. He has published extensively several accredited Computer vision and image Processing journals, and international and national conference proceedings. His research areas include computer vision, image processing, pattern recognition, and other image processing related areas, such as biometrics, medical imaging, and nuclear medicine.

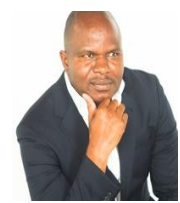

**Prof. Viriri** serves as a Reviewer for several accredited journals. He has served on program committees for numerous international and national conferences. He has graduated in several M.Sc and Ph.D. students.